\documentclass[letterpaper]{article} 
\usepackage[submission]{aaai23}  
\usepackage{times}  
\usepackage{helvet}  
\usepackage{courier}  
\usepackage[hyphens]{url}  
\usepackage{graphicx} 
\usepackage{natbib}  
\usepackage{caption} 
\usepackage{algorithm}
\usepackage{algorithmic}

\usepackage{amssymb}
\usepackage{pifont}
\newcommand{\cmark}{\ding{51}}%

\usepackage[graphicx]{realboxes}
\usepackage{multirow}
\usepackage{tabularx}
\usepackage{adjustbox}
\usepackage{bigstrut}
\usepackage{booktabs}
\usepackage{comment}

\usepackage{newfloat}
\usepackage{listings}
\DeclareCaptionStyle{ruled}{labelfont=normalfont,labelsep=colon,strut=off} 
\floatstyle{ruled}
\newfloat{listing}{tb}{lst}{}
\floatname{listing}{Listing}
\title{When to Use What: An In-Depth Comparative Empirical Analysis of OpenIE Systems for Downstream Applications}
\author{
    Kevin Pei\textsuperscript{\rm 1}, Ishan Jindal\textsuperscript{\rm 2}, Kevin Chen-Chuan Chang\textsuperscript{\rm 1}, Chengxiang Zhai\textsuperscript{\rm 1}, Yunyao Li\textsuperscript{\rm 3}\\
    {\normalfont \textsuperscript{\rm 1} Grainger College of Engineering, University of Illinois at Urbana-Champaign}\\
    {\normalfont \textsuperscript{\rm 2} IBM Research}
    {\normalfont \textsuperscript{\rm 3} Apple Knowledge Platform}\\
    {\normalfont\{kspei2,kcchang,czhai\}@illinois.edu, Ishan.Jindal@ibm.com, yunyaoli@apple.com}
}
\affiliations{
    \textsuperscript{\rm 1}Association for the Advancement of Artificial Intelligence\\

    1900 Embarcadero Road, Suite 101\\
    Palo Alto, California 94303-3310 USA\\
    publications23@aaai.org
}

\usepackage{bibentry}

\begin{document}

\maketitle

\begin{abstract}
Open Information Extraction (OpenIE) has been used in the pipelines of various NLP tasks.
Unfortunately, there is no clear consensus on which models to use in which tasks.
Muddying things further is the lack of comparisons that take differing training sets into account.
In this paper, we present an application-focused empirical survey of neural OpenIE models, training sets, and benchmarks in an effort to help users choose the most suitable OpenIE systems for their applications.
We find that the different assumptions made by different models and datasets have a statistically significant effect on performance, making it important to choose the most appropriate model for one's applications.
We demonstrate the applicability of our recommendations on a downstream Complex QA application.
\end{abstract}
\section{Introduction}

Open Information Extraction (OpenIE) is the task of extracting relation tuples from plain text \cite{angeli2015leveraging}. 
In its simplest form, OpenIE extracts information in the form of tuples consisting of \textit{subject\emph{(S)}}, \textit{predicate\emph{(P)}}, \textit{object\emph{(O)}}, and any \textit{additional arguments\emph{(A)}}. 
OpenIE is open domain, intended to be easy to deploy in different domains without fine-tuning. 
The tuples extracted by OpenIE consist of tokens from the original text, allowing for the extraction of all relations regardless of type. 
Models largely fall into two categories: models with hand-crafted extraction patterns and models with automatically learned extraction patterns \cite{niklaus2018survey}. 
The increasing availability of semi-automatically generated training datasets using previous OpenIE methods \cite{cui2018neural}, as well as significant advances in deep learning modeling techniques such as LSTM have led to the development of state-of-the-art neural models that automatically learn extraction patterns \cite{cui2018neural, garg2018supervising}.

Since its introduction in \citet{etzioni2008open}, OpenIE has attracted a large amount of attention by the research community for use as a tool for a wide range of downstream NLP tasks such as Slot Filling \cite{soderland2013open, angeli2015leveraging}, Question Answering (QA) \cite{fader2013paraphrase, khot2017answering}, Summarization \cite{cao2018faithful, ponza2018facts}, and Event Schema Induction \cite{balasubramanian2013generating, romadhony2019utilizing}.
However, there is no real consensus on which OpenIE model is best for which application.
We can observe this lack of consensus in summarization, where different papers use OLLIE \cite{christensen2014hierarchical}, MinIE \cite{ponza2018facts}, and Stanford CoreNLP \cite{cao2018faithful, zhang2021far} as their OpenIE models.
Different applications may also have different best OpenIE models.
As an example, choosing a model that assumes all relations only have a subject and object may not be suitable for event schema induction since that excludes any event schemas with more than two entities.
The papers that introduce new OpenIE models and datasets do not specify how downstream applications would be impacted by the different assumptions those papers make about which relations to extract.

We find that prior OpenIE surveys are also insufficient to find the best OpenIE model for a given application.
The only previous application-focused OpenIE survey we found was \citet{mausam2016open}.
However, this survey does not identify the desired properties of OpenIE for those applications or provide an empirical comparison of OpenIE systems.
\citet{niklaus2018survey} provide a taxonomy of OpenIE models, but does not include any comparison of datasets or an empirical study.
\citet{glauber2018systematic} and \citet{claro2019multilingual} also do not provide an empirical application-focused survey.

Another problem is the lack of apples-to-apples comparisons between OpenIE models.
Comparisons should keep the training set, benchmark, and evaluation metric constant when comparing models to eliminate confounders.
Unfortunately, the papers that introduce new OpenIE models and datasets often do not provide this apples-to-apples comparison.
For example, CopyAttention \cite{cui2018neural}, SpanOIE \cite{zhan2020span}, IMoJIE \cite{kolluru2020imojie}, and OpenIE6 \cite{kolluru2020openie6} all compare their model to models trained on different training sets.
OpenIE6 reports performance on WiRE57 which Multi$^2$OIE \cite{ro2020multi} does not, but Multi$^2$OIE reports performance on ReOIE2016 which OpenIE6 does not.
Because the training set can greatly affect the performance of a neural model, we focus on selecting both the appropriate OpenIE model and training set, which we refer to as an \textit{OpenIE System}. 

To resolve our lack of understanding, we focus on the particular question: \textit{How do I choose a particular OpenIE system for a given application?}
Different implicit assumptions about OpenIE may have a significant impact on the performance of downstream applications such as the assumptions that all relations are verb-based \cite{zhan2020span} or that all relations have only a subject and object \cite{kolluru2020imojie}. 
To answer this question an apples-to-apples comparison must be conducted for different application settings, keeping the training data, benchmark, and evaluation criteria constant while comparing models.

Because it is impractical to find the best model for every application given the many possible applications of OpenIE, we instead characterize applications based on what properties they desire from OpenIE.
For example, the desire for N-ary relation extraction by event schema induction.
We use these properties to characterize OpenIE models and datasets and then evaluate whether those properties result in meaningful differences in performance.
In the process of answering these questions we provide an extensive apples-to-apples comparison of existing neural OpenIE models such that a practitioner can utilize our practical observations to effectively select a neural OpenIE model and training set for their downstream application. 
Finally, we apply our recommendations to a downstream Complex QA task.
We hope our survey provides insight into how to select OpenIE models and datasets for future practitioners' applications. 
In summary, our contributions are as follows:
\begin{itemize}
    \item We propose a comprehensive taxonomy that covers OpenIE training sets, benchmarks, evaluation metrics, and neural models.
    \item We present an extensive empirical comparison of different models on different datasets with recommendations based on the results.
    \item We perform a case study on Complex QA to show the efficacy of our recommendations.
\end{itemize}

To the best of our knowledge, our survey is the only application-focused empirical survey on OpenIE datasets and recent neural OpenIE methods.

\begin{table*}[]
\centering
\resizebox{\textwidth}{!}{
\begin{tabular}{lccccc}
\toprule
\multicolumn{1}{l}{}                                       & \begin{tabular}[c]{@{}c@{}}Question \\ Answering\end{tabular} & Slot Filling          & \begin{tabular}[c]{@{}c@{}}Event Schema \\ Induction\end{tabular} & Summarization         & \begin{tabular}[c]{@{}c@{}}Knowledge Base \\ Population\end{tabular} \\ \midrule
\textbf{\textsl{HR}}: Higher Recall                & \cmark                                         & \cmark & \cmark                                             & \cmark & \cmark                                                                                               \\
\textbf{\textsl{HP}}: Higher Precision             &                                          & \cmark &                                              & \cmark &                                                                                                 \\
\textbf{\textsl{N-ary}}: N-ary Relation Extraction              & \cmark                                         & \cmark & \cmark                                             &  &                                                                                                 \\
\textbf{\textsl{IN}}: Inferred Relation Extraction     & \cmark                                         & \cmark & \cmark                                             & \cmark &                                                                                                 \\
\textbf{\textsl{FE}}: Fast Extraction &                                          &  &                                              & &  \cmark                                                                                               \\ \bottomrule
\end{tabular}%
}
\caption{Properties explicitly mentioned in application papers as motivation for choosing a particular OpenIE model or as a way to improve performance within a case study. There are additional desired properties we omit that no existing models or datasets have, such as the canonicalization of extracted relations and extracting relations from imperative sentences \cite{fader2013paraphrase, khot2017answering, zhang2021far}.}
\label{table:applications}
\end{table*}

\section{Motivating Applications}

In this section, we identify the properties of OpenIE systems desired by 5 downstream applications: \textit{Slot Filling}, \textit{Question Answering (QA)}, \textit{Summarization}, \textit{Event Schema Induction}, and \textit{Knowledge Base Population}.
We survey how OpenIE is used in each of the applications and the properties explicitly desired based on the corresponding papers, either as motivation for choosing a given OpenIE model or within a case study as a property that would improve performance.

The desired properties we observe are Higher Recall, Higher Precision, N-ary Relation Extraction, Inferred Relation Extraction, and Fast Extraction.
We define an "Inferred Relation" (IN) to be a relation where the predicate contains words that are not in the original sentence.
For example, given the sentence "Bill Gates, former CEO of Microsoft, is a Harvard dropout", the relation (Bill Gates, was, former CEO of Microsoft) can be inferred even though "was" is not in the original sentence.
We define an "N-ary Relation" (N-ary) to be a relation with more arguments than just (subject, predicate, object).
For instance, the relation (Alice, baked, Bob, a pie) has an additional argument.

\noindent\textbf{Slot Filling}
Slot filling is a task where an incomplete tuple must be completed using information from a given corpus \cite{chen2019bert}.
For example, given the incomplete tuple (Obama, born in, ?), extract (Obama, was born in, Honolulu) using information from the corpus.
In this task, OpenIE is used to extract complete tuples which are used to fill slots by linking those tuples to tuples missing slots using entity linking methods.
OpenIE can be used by extracting complete tuples which are linked to the incomplete tuple using entity linking methods.
The incomplete tuple can then be completed by using the complete tuple extracted using OpenIE.
Because OpenIE is not constrained to a predefined schema, it is useful for extracting different surface forms of relations.
\citet{soderland2013open}, \citet{angeli2015leveraging}, \citet{soderland2015university}, and \citet{soderland2015combining} take advantage of how correct relations often appear multiple times in text to match empty slots to the highest precision OpenIE tuple.
\citet{soderland2013open}, \citet{angeli2015leveraging}, \citet{soderland2015university}, and \citet{soderland2015combining} all state in their case studies they would benefit from the ability to extract inferred relations (\textsl{IN}), and \citet{soderland2015university} and \citet{soderland2015combining} state they would benefit from the ability to extract n-ary relations (\textsl{N-ary}).
These two properties allow more surface forms of relations to be extracted, which allows for more slots to be filled.

\noindent\textbf{Question Answering}
We focus on 2 subtasks of QA that utilize OpenIE: Open-domain Question Answering (OpenQA) and Complex QA.
\citet{fader2013paraphrase,fader2014open,yin2015answering}, and \citet{clark2018think} are OpenQA methods that use retrieval-based methods to match OpenIE extractions to questions.
OpenQA involves answering questions given a large database \cite{fader2014openqa}.
By rewriting queries into incomplete tuples, it is possible to use relations extracted from the database to answer queries by filling in the missing slots in the query.
For example, rewriting the query "Where was Obama born?" into slot filling the tuple (Obama, born in, ?) and answering using the relation (Obama, was born in, Honolulu).
\citet{fader2014open} and \citet{yin2015answering} obtain those relations from a knowledge base of OpenIE extractions.

Complex questions must use information from multiple sentences to find answers and require inferring relationships between multiple entities\cite{chali2009complex}.
\citet{khot2017answering} and \citet{lu2019answering} generate graphs from extracted relation tuples, then reason over these graphs to answer the questions.

In all QA applications surveyed, high recall (\textsl{HR}) is desired, with \citet{lu2019answering} using a custom OpenIE method specifically to obtain a higher recall.
\citet{yin2015answering}'s case studies state that \textsl{N-ary} in particular would benefit performance while \citet{lu2019answering} uses a custom OpenIE method that supports \textsl{IN}.
Several methods already paraphrase questions so that the surface forms of extracted relations match at least one of the question paraphrases, indicating that extracting more surface forms of a relation would answer more questions \cite{fader2013paraphrase, fader2014open, yin2015answering}.

\noindent\textbf{Summarization} 
OpenIE addresses the problems of redundancy and fact fabrication in summarization.
To combat redundancy in hierarchical and abstractive summarization, OpenIE is used to ensure that the generated summary does not have repeated relations or more relations than the gold standard summary \cite{christensen2014hierarchical, zhang2021far}.
To combat fact fabrication in abstractive summarization, OpenIE is used to ensure that the generated summary only contains relations from the original text \cite{cao2018faithful,zhang2021far}.
In summarization tasks, \textsl{HR} is useful to ensure summaries contain all information, with \citet{ponza2018facts} citing greater diversity of extractions as a way to improve performance.
The exception is \citet{zhang2021far}, where high precision (\textsl{HP}) is desired in order to reduce redundant extractions.

\noindent\textbf{Event Schema Induction} 
Event Schema Induction is the automatic discovery of patterns that indicate events and the agents and their roles within that event.
Extracted relation tuples can be used to find surface forms of events, with repeated extracted tuples being used to induce event schemas.
The open nature of OpenIE allows for events to be found regardless of the domain or surface form of the event.
Predicates of extracted tuples are generally mapped to events and arguments to agents of the event \cite{balasubramanian2013generating,romadhony2019utilizing,sahnoun2020event}.
\textsl{HR} is useful for Event Schema Induction for the same reason it is useful for Slot Filling: finding more surface forms of an event allows for more event schemas to be induced.
\citet{sahnoun2020event} also specifically desire \textsl{IN} so that more event schemas can be learned, while \citet{balasubramanian2013generating} state that \textsl{N-ary} would improve performance.

\noindent\textbf{Knowledge Base Population} 
OpenIE's open-domain nature has led to its use in automatically populating knowledge bases (KBs).
The relations extracted by OpenIE can be used to create new nodes and edges in KBs.
\citet{muhammad2020open} and \citet{kroll2021toolbox} use learning-based OpenIE models because of their ability to generalize to unseen relations and achieve \textsl{HR}.
\citet{kroll2021toolbox} also explicitly chooses Stanford CoreNLP and OpenIE6 because of their fast extraction times (\textsl{FE}).

We use these desired properties to formulate research questions in Section \ref{sub:rq}.
Table \ref{table:applications} provides a summary of applications and their explicitly desired properties.

\begin{table*}[]
\centering
\resizebox{\textwidth}{!}{%
\begin{tabular}{clccrrrrc}
\toprule
& \multicolumn{1}{l}{\textbf{Dataset}}             & \textbf{\begin{tabular}[c]{@{}c@{}}Creation Method\end{tabular}} & \textbf{Source}                                                          & \multicolumn{1}{c}{\textbf{\#Extractions}} & \multicolumn{1}{c}{\textbf{\#IN}} & \multicolumn{1}{c}{\textbf{\#N-ary}} \\ \midrule
\multirow{5}{*}{\begin{tabular}[c]{@{}c@{}}Training\\ Sets\end{tabular}} &SpanOIE                                          & \begin{tabular}[c]{@{}c@{}}Weak Labeling\end{tabular}            & Wikipedia                                                                & 2,175K                                     & 2K                                                                       & 231K                                 \\
& OIE4                                             & \begin{tabular}[c]{@{}c@{}}Weak Labeling\end{tabular}            & Wikipedia                                                                & 181K                                       & 3K                                                                        & 34K                                  \\
& IMoJIE                                           & \begin{tabular}[c]{@{}c@{}}Weak Labeling\end{tabular}            & Wikipedia                                                                & 215K                                       & 3K                                                                        & 0                                    \\
& LSOIE                                            & \begin{tabular}[c]{@{}c@{}}Weak Labeling\end{tabular}  & \begin{tabular}[c]{@{}c@{}}QA-SRL 2.0 Wikipedia, Science\end{tabular} & 101K                                       & 0                                                                          & 32K                                  \\ \midrule
\multirow{6}{*}{\begin{tabular}[c]{@{}c@{}}Test\\ Sets\end{tabular}}  & OIE2016                                          & \begin{tabular}[c]{@{}c@{}}Weak Labeling\end{tabular}            & QA-SRL                                                                   & 1,730                                      & 359                                                                       & 708                                  \\
& WiRe57                                           & \begin{tabular}[c]{@{}c@{}}Manual Annotation\end{tabular}        & \begin{tabular}[c]{@{}c@{}}Wikipedia and Newswire\end{tabular}        & 343                                        & 173                                                                       & 79                                   \\
& ReOIE2016                                        & \begin{tabular}[c]{@{}c@{}}Manual Annotation\end{tabular}        & OIE2016                                                                  & 1,508                                      & 155                                                                      & 611                                  \\
& CaRB                                             & \begin{tabular}[c]{@{}c@{}}Crowdsourced Annotation\end{tabular}  & OIE2016                                                                  & 5,263                                      & 736                                                                       & 683                                  \\
& LSOIE                                            & \begin{tabular}[c]{@{}c@{}}Weak Labeling\end{tabular}            & \begin{tabular}[c]{@{}c@{}}QA-SRL 2.0 Wikipedia, Science\end{tabular} & 22,376                                     & 0                                                                      & 4,920                               \\ \bottomrule
\end{tabular}%
}
\caption{Comparison of the attributes of different datasets. $\#$Extractions: Number of Extractions, $\#$IN  : Number of inferred relations, $\#$N-ary: Number of N-ary Relations.}
\label{table:datasets}
\end{table*}

\section{Datasets}

In this section, we discuss the differences between different OpenIE training sets and benchmarks and what properties they possess.
We also briefly touch on different evaluation metrics.
The common principles that guide the creation of OpenIE training sets and benchmarks are \textit{Assertedness}, \textit{Minimal Propositions/Atomicity}, and \textit{Completeness and Open Lexicon} \cite{stanovsky2016creating, lechelle2018wire57, bhardwaj2019carb}.
\textit{Assertedness} means the relation must be implied by the original sentence alone.
\textit{Minimal Propositions} means each relation should include as few words as possible while retaining meaning.
\textit{Completeness and Open Lexicon} means all relations should be extracted, without a predefined domain or scope.

Despite these common guiding principles, OpenIE datasets still differ in their creation method and subsequently have differing properties.
We provide statistics about different datasets in table \ref{table:datasets}.

\subsection{Training Datasets}

Given how data-hungry deep learning models are and how costly it is to manually label OpenIE datasets, most OpenIE training sets are weakly labeled using high confidence extractions from prior OpenIE models to get "silver-standard" labels.
\textbf{CopyAttention} \cite{cui2018neural}, \textbf{SpanOIE} \cite{zhan2020span}, and \textbf{OIE4} \cite{kolluru2020imojie} are training sets consisting of high confidence OpenIE4 extractions from Wikipedia.
Unlike CopyAttention and OIE4, SpanOIE includes low-quality extractions with pronoun arguments.
The~\textbf{IMoJIE}~dataset \cite{kolluru2020imojie} attempts to get higher quality labels by combining Wikipedia extractions from OpenIE4, ClausIE, and RNNOIE, using a common scoring metric to combine extractions and filtering out repeated extractions.
The \textbf{LSOIE} training set \cite{solawetz2021lsoie} is composed of automatically converted Semantic Role Labeling (SRL) extractions with high inter-annotator agreement from the Wikipedia and Science domain of the crowdsourced QA-SRL Bank 2.0 dataset.
Because this dataset is derived from SRL, all relations are assumed to be verb-based and only contain words in the original sentence.

\subsection{Benchmarks}

\textbf{OIE2016} \cite{stanovsky2016creating} is a benchmark for OpenIE automatically derived from the crowdsourced QA-SRL dataset annotated on PropBank and Wikipedia sentences.
\textbf{WiRe57} \cite{lechelle2018wire57} consists of expert annotations for 57 sentences.
\textbf{CaRB} \cite{bhardwaj2019carb} uses crowdsourcing to re-annotate the sentences in the OIE2016 benchmark.
In contrast to other OpenIE datasets, each extraction contains as much information as possible in the arguments, meaning prepositions are part of the arguments and not the predicate.
\textbf{ReOIE2016} \cite{zhan2020span} uses manual annotation to re-annotate OIE2016 to attempt to resolve problems arising from incorrect extraction.
The \textbf{LSOIE} \cite{solawetz2021lsoie} training set has a corresponding benchmark derived using the same source and rules.
\textbf{BenchIE} \cite{gashteovski2021benchie} is derived from CaRB and is based on the idea that extracted relations need to exactly match at least one gold standard out of a set of equivalent manually annotated relations to be useful for downstream applications.


Some of these benchmarks introduce new evaluation metrics.
\textbf{OIE2016} introduces \textit{lexical matching}, which treats extraction as a binary classification task.
A predicted relation is matched to a gold standard relation if the heads of the predicate and all arguments are the same.
\textbf{WiRe57} and \textbf{CaRB} use \textit{word-level matching}, which calculate recall and precision based on the proportion of matching tokens in the predicted and gold standard relations.
The difference between the two is that there is a greater penalty to recall for \textbf{WiRe57} if there are fewer predicted relations than relations in the gold standard.


\textbf{BenchIE} uses sentence-level matching.
Sentence-level matching requires an exact match of the predicate and arguments instead of just the heads like lexical matching.
Instead of a set of gold tuples to represent the gold standard relations, fact sets represent the gold standard relations.
Each fact set consists of a set of equivalent relations and a predicted relation is considered a true positive for a given relation if it matches a relation within the fact set exactly.

Because of BenchIE's reliance on fact sets which other benchmarks lack, the BenchIE metric is only compatible with BenchIE and no other metrics can be used with the BenchIE dataset.
As a result, an apples-to-apples comparison of the BenchIE dataset and metric are not possible like with other datasets and metrics, so we do not report performance on BenchIE.

\begin{table}[]
\centering
\begin{adjustbox}{width=0.9\linewidth}
\begin{tabular}{lccc}
\toprule
\multicolumn{1}{l}{\textbf{Model}} & \textbf{\begin{tabular}[c]{@{}c@{}}Problem Formulation\end{tabular}} & \textbf{N-ary}    & \textbf{IN} \\ \midrule
SpanOIE                            & \begin{tabular}[c]{@{}c@{}}Labeling\end{tabular}          & \cmark &  \\
IMoJIE                             & Generation                                                              &  &  \\
Multi$^2$OIE                          & Labeling                                                                & \cmark &  \\
IGL-OIE                            & Labeling                                                                &  & \cmark \\
CIGL-OIE                           & Labeling                                                                &  & \cmark \\
OpenIE6                            & Labeling                                                                &  & \cmark \\ \bottomrule
\end{tabular}%
\end{adjustbox}
\caption{Comparison of neural OpenIE models.}
\label{table:models}
\end{table}

\section{Models}

In this section, we present recent neural OpenIE models and the properties that set them apart.
OpenIE models can be categorized based on how they formulate the OpenIE problem: as a text generation or labeling problem.
Different models also possess different properties depending on what assumptions they make about relations.
We provide overviews of the models in table \ref{table:models}.

\subsection{Generative Problem Formulation}

Generative OpenIE models cast OpenIE as a sequence-to-sequence problem, taking the sentence as input and attempting to generate all relations in the sentence as output.
\textbf{CopyAttention} \cite{cui2018neural} generates extractions using GloVe embeddings and a 3-layer stacked Long Short-Term Memory (LSTM) as the encoder and decoder.
\textbf{IMoJIE} \cite{kolluru2020imojie} builds upon CopyAttention by using BERT embeddings and introducing \textit{iterative extraction} to combat repeated extractions.
\textit{Iterative extraction} is the appending of extractions to the end of the sentence before being used as input so the model can identify what relations have previously been extracted at the cost of significantly reduced extraction speed.
\textbf{Adversarial-OIE} \cite{han2021generative} uses a Generative Adversarial Network (GAN) to generate extractions.
The model consists of a sequence generator model that takes BERT and position embeddings as input and generates extractions, and an adversary model that tries to distinguish between relations in the gold standard and relations extracted by the sequence generator model.

Generative models rely on a copy mechanism to copy vocabulary from the original sentence, meaning they can not generate tokens that are not in the original sentence and subsequently can not extract \textsl{IN} relations.

\subsection{Labeling Problem Formulation}

Labeling OpenIE models cast OpenIE as a sequence labeling problem, taking the sentence as input and labeling each token in the sentence with its role in each relation, usually using a BIO tagging scheme.
Labeling models can be further subdivided into Piecewise and Holistic Labeling.

\textbf{Piecewise Labeling} models label predicates and arguments in different stages.
\textbf{RnnOIE} \cite{stanovsky2018supervised} is a bi-directional LSTM (BiLSTM) transducer inspired by SRL that firsts labels predicates and then labels arguments with BIO tags for each extracted predicate.
\textbf{SpanOIE} \cite{zhan2020span} is also based on SRL, using a BiLSTM to perform span classification instead of BIO tagging.
Span classification enables the use of span features, which can be richer than word-level features.
\textbf{Multi$^2$OIE}'s \cite{ro2020multi} novelty is multi-head attention and BERT embeddings.
After labeling the predicates, multi-head attention is used between the predicate and the rest of the sentence to label the arguments.
\textbf{MILIE} \cite{kotnis2021integrating} introduces its own \textit{iterative prediction}, the process of extracting one component of the relation tuple at a time, for multilingual OpenIE.
Extraction can be performed predicate first, subject first, or object first, followed by any of the remaining components.
The intention was to make multilingual extraction easier, in case other languages benefited from different extraction orders.
\textbf{DetIE} \cite{vasilkovsky2022detie} uses ideas from single-shot object detection to make predictions more quickly than previous methods.

Uniquely, piecewise labeling models label all predicates in a sentence simultaneously and assume that for each predicate, there is only one set of arguments.
This means that they can not extract multiple relations that share the same predicate, unlike generative and holistic labeling models.

\textbf{Holistic Labeling} models label predicates and arguments simultaneously
\textbf{OpenIE6} \cite{kolluru2020openie6} introduces grid labeling, constraint rules, and conjunction rules to improve the labeling process.
Grid labeling is simultaneous extraction of multiple relations from a sentence.
Constraint rules are used to penalize certain things like repeated extractions or not extracting a relation for a head verb.
Conjunction rules are used to split relations containing conjunctions into two separate relations.
IGL-OIE is the first stage, using only grid labeling; CIGL-OIE is the second stage, adding in constraint rules; OpenIE6 is the final stage, using conjunction rules to handle relations that contain conjunctions.

Labeling models generally can not label tokens that are not in the original sentence, meaning they can not extract \textsl{IN} relations.
The exceptions are IGL-OIE, CIGL-OIE, and OpenIE6, which explicitly allows for the extraction of "be" relations even if they are not in the original sentence.

\section{Experiments}

In this section, we describe how we compare OpenIE models and datasets for the sake of recommendation.
In our experiments, we focus on the reported state-of-the art models for English OpenIE, the most widely-used English benchmarks with associated papers, and the 4 training sets used by those models.
We then use OpenIE in a downstream Complex QA task to demonstrate the applicability of our recommendations.

When comparing OpenIE systems, we place a greater emphasis on F1 score than AUC.
The original implementations of CaRB, OIE2016, and WiRe57 use the trapezoidal rule to calculate the area under the PR curve.
It is assumed that all PR curves have data points at recall 0, precision 1.
This means that methods without low recall points on the PR curve will have inflated AUC values.
For example, for the CaRB test set using the CaRB metric, CIGL-OIE trained on OIE4 has a minimum recall of 0.214 with precision 0.678 at confidence 1.0, while Multi$^2$OIE trained on SpanOIE has a minimum recall of 0.0003 with precision 0.5 at confidence 1.0.
Considering just these lowest recall data points, CIGL-OIE has an AUC of 0.180, while Multi$^2$OIE has an AUC of 0.0002, meaning CIGL-OIE already has a large inherent advantage in AUC without considering the higher confidence data points.
As a result, we consider the highest F1 score on the PR curve to be a better metric than AUC when evaluating overall model performance.

\subsection{Research Questions}
\label{sub:rq}

To find the best system for different applications, we test whether the properties of OpenIE models and datasets have a statistically significant effect on accuracy in benchmarks with properties that correspond to the desires of downstream applications.
If there is no significant difference between the performance of OpenIE systems with different properties, then the chosen system should be the system with the best overall empirical performance.
In addition to how model and dataset properties affect accuracy, we are also interested in how the choice of model affects efficiency in order to satisfy the fast extraction property (\textsl{FE}).
Generally, generative models are much slower than labeling models \cite{kolluru2020openie6}.
Subsequently, we investigate whether the efficiency difference between model types is significant.

\begin{enumerate}
        \item \textbf{R1:} How does whether a model supports N-ary relation (\textsl{N-ary}) extraction and whether the training set contains \textsl{N-ary} affect the F1 score of a model on test sets with \textsl{N-ary}?
    
        \item \textbf{R2:} How does whether a model supports inferred relation  (\textsl{IN}) extraction and whether the training set contains \textsl{IN} affect the F1 score of a model on test sets with or without \textsl{IN}?
    
        \item \textbf{R3:} How does the model type affect efficiency as measured by the number of sentences relations are extracted from per second (Sen./Sec)?
    
\end{enumerate}

\begin{table}[]
\centering
\begin{adjustbox}{width=\linewidth}
\begin{tabular}{lrrrrrrr}
\toprule
\multirow{2}{*}{Model} & \multirow{2}{*}{Sen./Sec.}      & \multicolumn{3}{c}{CaRB}                                               & \multicolumn{3}{c}{WiRE 57}                                            \\ \cmidrule(lr){3-5} \cmidrule(lr){6-8}
                                &                                 & \multicolumn{1}{c}{P} & \multicolumn{1}{c}{R} & \multicolumn{1}{c}{F1} & \multicolumn{1}{c}{P} & \multicolumn{1}{c}{R} & \multicolumn{1}{c}{F1} \\ \midrule
SpanOIE                         & 13.40                           & 0.474                 & 0.464                 & 0.433                  & 0.474                 & 0.374                 & 0.375                  \\
IMoJIE                          & 2.07                            & 0.598                 & 0.431                 & 0.488                  & 0.598                 & 0.355                 & 0.428                  \\
Multi$^2$OIE                    & 29.22                           & \textbf{0.626}        & 0.501                 & \textbf{0.552}         & \textbf{0.624}        & 0.419                 & \textbf{0.488}         \\
IGL-OIE                         & \textbf{84.07} & 0.574                 & 0.442                 & 0.497                  & 0.574                 & 0.365                 & 0.434                  \\
CIGL-OIE                        & 68.80                           & 0.490                 & \textbf{0.531}        & 0.503                  & 0.489                 & 0.429                 & 0.442                  \\
OpenIE6                         & 28.36                           & 0.394                 & 0.518                 & 0.438                  & 0.394                 & \textbf{0.463}        & 0.413                  \\ \hline
\end{tabular}%
\end{adjustbox}
\caption{Performances of different models with different evaluation metrics averaged across training and test data.}
\label{table:avg_model_performance}
\end{table}

\begin{table}[]
\centering
\begin{adjustbox}{width=\linewidth}
\begin{tabular}{llrrrrrr}
\toprule
\multirow{2}{*}{Training Set} & \multirow{2}{*}{Test Set} & \multicolumn{3}{c}{CaRB}                                               & \multicolumn{3}{c}{WiRE 57}                                            \\  \cmidrule(lr){3-5} \cmidrule(lr){6-8}
                                       &                           & \multicolumn{1}{c}{P} & \multicolumn{1}{c}{R} & \multicolumn{1}{c}{F1} & \multicolumn{1}{c}{P} & \multicolumn{1}{c}{R} & \multicolumn{1}{c}{F1} \\ \midrule
SpanOIE                                & OIE2016                   & 0.485          & 0.502                 & 0.478          & 0.484          & 0.420                 & 0.434          \\
OIE4                                   & OIE2016                   & 0.537          & 0.493                 & 0.511          & 0.536          & 0.410                 & 0.461          \\
LSOIE                                  & OIE2016                   & \textbf{0.620} & \textbf{0.538}        & \textbf{0.564} & \textbf{0.620} & \textbf{0.447}        & \textbf{0.508} \\
IMoJIE                                 & OIE2016                   & 0.446          & 0.449                 & 0.423          & 0.445          & 0.378                 & 0.382          \\ \midrule
SpanOIE                                & WiRe57                    & 0.411          & 0.371                 & 0.381          & 0.415          & 0.199                 & 0.261          \\
OIE4                                   & WiRe57                    & \textbf{0.470} & \textbf{0.374}        & \textbf{0.416} & \textbf{0.469} & 0.211                 & \textbf{0.289} \\
LSOIE                                  & WiRe57                    & 0.347          & 0.208                 & 0.257          & 0.347          & 0.127                 & 0.183          \\
IMoJIE                                 & WiRe57                    & 0.421          & 0.370                 & 0.373          & 0.419          & \textbf{0.223}        & 0.266          \\ \midrule
SpanOIE                                & ReOIE2016                 & 0.636          & \textbf{0.635}        & \textbf{0.615} & 0.636          & \textbf{0.622}        & \textbf{0.610} \\
OIE4                                   & ReOIE2016                 & \textbf{0.720} & 0.565                 & 0.597          & \textbf{0.720} & 0.553                 & 0.590          \\
LSOIE                                  & ReOIE2016                 & 0.622          & 0.527                 & 0.558          & 0.622          & 0.516                 & 0.552          \\
IMoJIE                                 & ReOIE2016                 & 0.586          & 0.588                 & 0.554          & 0.584          & 0.569                 & 0.544          \\ \midrule
SpanOIE                                & CaRB                      & 0.529          & 0.446                 & 0.471          & 0.526          & 0.313                 & 0.378          \\
OIE4                                   & CaRB                      & \textbf{0.604} & \textbf{0.449}        & \textbf{0.513} & \textbf{0.604} & \textbf{0.317}        & \textbf{0.412} \\
LSOIE                                  & CaRB                      & 0.532          & 0.344                 & 0.412          & 0.532          & 0.255                 & 0.337          \\
IMoJIE                                 & CaRB                      & 0.517          & 0.429                 & 0.447          & 0.514          & 0.315                 & 0.360          \\ \midrule
SpanOIE                                & LSOIE                     & 0.461          & 0.573                 & 0.501          & 0.461          & 0.525                 & 0.477          \\
OIE4                                   & LSOIE                     & 0.505          & 0.566                 & 0.532          & 0.505          & 0.517                 & 0.506          \\
LSOIE                                  & LSOIE                     & \textbf{0.653} & \textbf{0.686}        & \textbf{0.660} & \textbf{0.653} & \textbf{0.629}        & \textbf{0.629} \\
IMoJIE                                 & LSOIE                     & 0.419          & 0.508                 & 0.440          & 0.419          & 0.473                 & 0.427          \\ \hline
\end{tabular}%
\end{adjustbox}
\caption{Performances of different training and test sets averaged across models.}
\label{table:avg_dataset_performance}
\end{table}

\subsection{Experimental Setup}

\noindent\textbf{Models:} We compare \textit{SpanOIE}, \textit{IMoJIE}, \textit{Multi$^2$OIE}, and the 3 stages of OpenIE6: \textit{IGL-OIE}, \textit{CIGL-OIE}, and \textit{OpenIE6}. 
These models are publicly available, with Multi$^2$OIE and OpenIE6 being state-of-the-art for labeling models and IMoJIE being state-of-the-art for generative models. 
For each model, we train them with their original dev set and their original hyperparameters.
We run all experiments using a Quadro RTX 5000 GPU.

\noindent\textbf{Training Datasets:} We train the models on the \textit{SpanOIE}, \textit{OIE4}, \textit{IMoJIE}, and \textit{LSOIE} training sets. 
The LSOIE training set contains both the Science and Wikipedia domain sentences to increase the amount of training data. 
Due to the input structure of SpanOIE and Multi$^2$OIE models, they can not be trained on training datasets with inferred relations. 
We remove any inferred relations from the training sets of those models.
Similarly, as IMoJIE and OpenIE6 can not extract N-ary relations, we convert all N-ary relations in the training set into binary relations by moving arguments beyond the subject and object into the object. 
For instance, the relation (Alice, baked, Bob, a pie) was converted into (Alice, baked, Bob a pie). 
None of these limitations apply to the test sets.

\noindent\textbf{Benchmarks:} We evaluate all the models on the publicly available English benchmarks \textit{OIE2016}, \textit{WiRE57}, \textit{ReOIE2016}, \textit{CaRB}, and \textit{LSOIE}. 

\noindent\textbf{Evaluation Metrics:} We use \textit{OIE2016}'s, \textit{WiRE57}'s, and \textit{CaRB}'s metrics for evaluation.
We compare performance using primarily F1 score to address \textsl{HR} and \textsl{HP} and sentences extracted per second to address \textsl{FE}.
We perform student's t-test between OpenIE system, test set, and evaluation metric configurations to answer \textbf{R1}, \textbf{R2}, and \textbf{R3}. 
For \textbf{R1} and \textbf{R2} the t-scores are computed using the per-sentence F1 scores of each method.
For \textbf{R3} the t-scores are computed using the mean sentences per second for each training set and test set combination for a given model, resulting in 24 data points per model.
We do not measure the time required to extract from each individual sentence because it can be difficult to measure given the parallel execution of each mode.

\section{Results}

\subsection{Overall Evaluation}
In this section, we perform an apples-to-apples comparison among different OpenIE systems to first determine the SoTA OpenIE model and then to determine the best general-purpose OpenIE training dataset. 

\noindent\textbf{Best OpenIE Model} We compare the different models on different evaluation metrics averaged across different training and test sets in Table \ref{table:avg_model_performance}. 
We observe that across all evaluation metrics  Multi$^2$OIE and  the CIGL-OIE have the highest or second highest F1 score.
This means that independent of training and test sets the \textit{Labeling} OpenIE models are better than the \textit{Span Classification} and \textit{Generative} models. 
We also observe that \textit{Labeling} OpenIE models are more efficient than the \textit{Span Classification} and \textit{Generative} models, extracting from more sentences per second.

\noindent\textbf{Best OpenIE Training Set} Because performance on a test set is also greatly dependent on the training set depending on the domain and generation method of the training and test sets, we determine the best training set for each test set.
In Table \ref{table:avg_dataset_performance}, we compare different training and test set combinations with different evaluation metrics averaged across models.
We observe that the models trained on LSOIE training set perform best on the OIE2016 and LSOIE test sets. 
This is because the LSOIE training set and the OIE2016 and LSOIE test sets are derived from different versions of QA-SRL and generated using the same rules.  
On the WiRe57, ReOIE2016, and CaRB test sets, we observe that the models trained on the OIE4 and SpanOIE training sets generally perform the best. 
It is likely because the OIE4 and SpanOIE training sets contain \textsl{N-ary} and \textsl{IN} relations like the WiRe57, ReOIE2016, and CaRB test sets.
Other training sets lack one or both of these properties.

\begin{table}[]
\centering
\begin{adjustbox}{width=\linewidth}
\begin{tabular}{l|l}
\toprule
\textbf{Sentence}  & \begin{tabular}[c]{@{}l@{}}A short distance to the east, NC 111 diverges \\ on Greenwood Boulevard.\end{tabular}   \\ \hline
\textbf{Multi2OIE} & (NC 111, diverges, on Greenwood Boulevard)                                                                         \\ \hline
\textbf{CIGL-OIE}  & \begin{tabular}[c]{@{}l@{}}(NC 111, diverges, A short distance to the east \\ on Greenwood Boulevard)\end{tabular} \\ \midrule
\end{tabular}%
\end{adjustbox}
\caption{A demonstration that CIGL-OIE tends to extract longer arguments than Multi$^2$OIE. Multi$^2$OIE is trained on SpanOIE and CIGL-OIE is trained on OIE4. The sentence is from the CaRB test set.}
\label{table:extraction_comparison}
\end{table}

Of the two models with the highest average CaRB F1 scores, Multi$^2$OIE and CIGL-OIE, Multi$^2$OIE has higher average precision while CIGL-OIE has higher average recall.
CIGL-OIE tends to extract longer objects than Multi$^2$OIE as seen in table \ref{table:extraction_comparison}, which may explain this difference.

\subsection{Research Questions}

To answer our research questions, we perform student's t-test using the CaRB F1 scores of the highest scoring model, training set, and test set combinations for each setting.
We perform comparisons of OpenIE systems, where one aspect (model or training set) is changed and the other aspects are kept constant.
Then, we choose the test set and evaluation metric for the two settings that results in the highest t-score between methods, which is used to answer \textbf{R1} and \textbf{R2}. 

For \textbf{R1}, we conclude (1) regardless of training set, the best \textsl{N-ary} models perform better than the best non-\textsl{N-ary} models; (2) regardless of the model, training on the best \textsl{N-ary} training sets results in higher performance than training on the best non-\textsl{N-ary} training sets. 
Therefore \textbf{if an application benefits from \textsl{N-ary}, then the best OpenIE system should include either a \textsl{N-ary} model, \textsl{N-ary} training set, or both}, with both being preferred if both are available.

For \textbf{R2}, we infer that \textsl{IN} models are better than non-\textsl{IN} models when there is either a \textsl{IN} training and \textsl{IN} test set, or when there is a non-\textsl{IN} training and non-\textsl{IN} test set.
\textsl{IN} training sets are better than non-\textsl{IN} training sets when there is an \textsl{IN} model and \textsl{IN} test set.
In the case of non-\textsl{IN} and \textsl{IN} test set, it is unclear whether \textsl{IN} or non-\textsl{IN} training sets are superior.
Therefore \textbf{if an application benefits from \textsl{IN}, then the chosen training set and model should either both be \textsl{IN} or both be non-\textsl{IN}. 
If an application benefits from non-\textsl{IN}, then the chosen training set should be non-\textsl{IN}, and subsequently the chosen model should be \textsl{IN}.}

For \textbf{R3}, we compare the efficiency of the sole generative model, IMoJIE, to the efficiency of every other model.
From our results, we observe that every other model is faster than IMoJIE and the difference is statistically significant. 
This makes intuitive sense, since generative models tend to have longer execution times than labeling models and it has been previously shown that IMoJIE is exceptionally slow compared to other OpenIE methods \cite{kolluru2020openie6}.
Therefore \textbf{if an application is concerned about efficiency, then the chosen OpenIE model should not be a generative model.}

\section{A Case Study: Complex QA}

To verify our recommendations, we perform a case study using QUEST \cite{lu2019answering}, a Complex QA method that uses OpenIE to extract entities and predicates from the question and from answer documents to generate knowledge graphs.
The nodes are entities derived from the subjects and objects, while the edges are predicates.
The knowledge graph is matched to the entities in the question and traversed to find potential answers.
Because more extractions result in a larger knowledge graph, QUEST benefits from \textsl{HR} which the authors use their own rule-based OpenIE method to achieve.

\subsection{Experimental Setup}

To test our recommendations, we replace the OpenIE method used by the authors with Multi$^2$OIE trained on SpanOIE, CIGL-OIE trained on OIE4, and OpenIE6 trained on OIE4.
We chose these models and training sets because they are the ones with the overall highest CaRB recall and F1 scores that match the properties desired by complex QA, namely \textsl{N-ary} and \textsl{IN}.

One caveat is that in order for QUEST to connect entities from multiple sentences, they must have the same surface form.
Because OpenIE methods often extract long subjects and objects that include adjectives and modifiers, if the subject or object of an extraction contains entities extracted by QUEST, we add additional relations using those entities.
For example, in the sentence "Hector Elizondo was nominated for a Golden Globe for his role in Pretty Woman," QUEST may extract the entities "Hector Elizondo," "Golden Globe," and "Pretty Woman."
If an OpenIE method were to extract the triple \textit{("Hector Elizondo", "was nominated", "for a Golden Globe for his role in Pretty Woman")}, we would add the additional extractions \textit{("Hector Elizondo", "was nominated", "Golden Globe")} and \textit{("Hector Elizondo", "was nominated", "Pretty Woman")}.
QUEST also performs preprocessing before running its rule-based OpenIE method, most prominently replacing pronouns with the entities they refer to.
This is because nodes in the knowledge graph can not be made using pronouns.
We replace pronouns using the same method QUEST does before running any OpenIE method.

We run QUEST using the CQ-W question set and search for answers in the Top-10 Google document set used in their paper. 
Because CIGL-OIE has the highest CaRB recall and OpenIE6 has the highest WiRe57 recall, we expect that using either of them will result in higher downstream performance than using Multi$^2$OIE.

\begin{table}[]
\centering
\begin{adjustbox}{width=\linewidth}
\begin{tabular}{lllrrr}
\toprule
\multicolumn{1}{l}{OpenIE} & \multicolumn{1}{c}{Questions} & \multicolumn{1}{c}{Documents} & \multicolumn{1}{c}{MRR} & \multicolumn{1}{c}{P@1} & \multicolumn{1}{c}{Hit@5} \\ \midrule
QUEST                                      & CQ-W                                   & Top 10                                 & 0.132                            & 0.080                            & 0.167                              \\ \midrule
CIGL-OIE                                   & CQ-W                                   & Top 10                                 & \textbf{0.111}                           & \textbf{0.060}                            & \textbf{0.167}                              \\
OpenIE6                                    & CQ-W                                   & Top 10                                 & 0.104                            & 0.060                            & 0.147                              \\
Multi2OIE                                  & CQ-W                                   & Top 10                                 & 0.094                            & 0.053                            & 0.140                              \\ \bottomrule
\end{tabular}%
\end{adjustbox}
\caption{Performance of QUEST using different OpenIE methods on the CQ-W dataset using the Top 10 Google documents. }
\label{table:QUEST}
\end{table}

\subsection{Evaluation}

We compare the Mean Reciprocal Rank (MRR), Precision@1 (P@1), and Hit@5 for each OpenIE model.
The results of our case study are summarized in table \ref{table:QUEST}.
We observe higher performance of CIGL-OIE and OpenIE6 than Multi$^2$OIE on QUEST, which matches our expectations based on the higher recall of CIGL-OIE and OpenIE6.
Our case study demonstrates the applicability of our empirical study to the use of OpenIE methods in downstream applications.

An important note is that oftentimes a great deal of pre- and post-processing is necessary to adapt OpenIE for different downstream applications.
Removing pronouns and adding additional entity-based extractions was necessary to achieve reasonable performance with different OpenIE methods in QUEST.
Even after modifying Multi$^2$OIE, CIGL-OIE, and OpenIE6 in this way, their performance is less than the original performance of QUEST.
As a result, it is important for practitioners to not just consider the performance of OpenIE models on test sets matching their application, but to also consider how to adapt OpenIE to the specific needs of their application.

\section{Conclusion}

In this paper, we presented an application-focused empirical comparison of recent neural OpenIE models, training sets, and benchmarks.
Our experiments showed that the different properties of OpenIE models and datasets affect the performance, meaning it is important to choose the appropriate system for a given application and not just choose whatever model is state-of-the-art.
We hope that this survey helps users identify the best OpenIE system for their downstream applications and inspires new OpenIE research into addressing the properties desired by downstream applications.

\bibliography{aaai23}

\appendix

\label{sec:appendix}

\section{Appendix A: Empirical Results}
\label{sec:AppendixB}

In this section, we show the empirical results of training each model on a variety of training sets and evaluating them on a variety of test sets with different evaluation metrics. 
We also show the empirical results of our student's t-tests comparing different OpenIE systems.

\begin{table*}[]
\centering
\resizebox{\textwidth}{!}{%
\begin{tabular}{lllrrrrrrrrrrrrr}
\toprule
\multicolumn{1}{c}{\multirow{2}{*}{Model}} & \multicolumn{1}{c}{\multirow{2}{*}{Training set}} & \multicolumn{1}{c}{\multirow{2}{*}{Test set}} & \multicolumn{1}{c}{\multirow{2}{*}{Sen./Sec}} & \multicolumn{4}{c}{OIE2016}                                                                      & \multicolumn{4}{c}{WiRe57}                                                                       & \multicolumn{4}{c}{CaRB}                                                                         \\ \cmidrule(lr){5-8}  \cmidrule(lr){9-12}  \cmidrule(lr){13-16} 
\multicolumn{1}{c}{}                       & \multicolumn{1}{c}{}                              & \multicolumn{1}{c}{}                          & \multicolumn{1}{c}{}                          & \multicolumn{1}{c}{P} & \multicolumn{1}{c}{R} & \multicolumn{1}{c}{F1} & \multicolumn{1}{c}{AUC} & \multicolumn{1}{c}{P} & \multicolumn{1}{c}{R} & \multicolumn{1}{c}{F1} & \multicolumn{1}{c}{AUC} & \multicolumn{1}{c}{P} & \multicolumn{1}{c}{R} & \multicolumn{1}{c}{F1} & \multicolumn{1}{c}{AUC} \\ \midrule
SpanOIE                                    & SpanOIE                                           & OIE2016                                       & 16.65                                         & 0.704                 & 0.792                 & 0.745                  & 0.675                   & 0.576        & 0.376                 & 0.455                  & 0.296          & 0.576        & 0.459                 & 0.511                  & 0.362          \\
\bottomrule
\end{tabular}%
}
\caption{A table that lists performances of different OpenIE systems on the OIE2016 benchmark.}
\end{table*}

\begin{table*}[]
\centering
\resizebox{\textwidth}{!}{%
\begin{tabular}{lllrrrrrrrrrrrrr}
\toprule
\multicolumn{1}{c}{\multirow{2}{*}{Model}} & \multicolumn{1}{c}{\multirow{2}{*}{Training set}} & \multicolumn{1}{c}{\multirow{2}{*}{Test set}} & \multicolumn{1}{c}{\multirow{2}{*}{Sen./Sec}} & \multicolumn{4}{c}{OIE2016}                                                                      & \multicolumn{4}{c}{WiRe57}                                                                       & \multicolumn{4}{c}{CaRB}                                                                         \\ \cmidrule(lr){5-8}  \cmidrule(lr){9-12}  \cmidrule(lr){13-16} 
\multicolumn{1}{c}{}                       & \multicolumn{1}{c}{}                              & \multicolumn{1}{c}{}                          & \multicolumn{1}{c}{}                          & \multicolumn{1}{c}{P} & \multicolumn{1}{c}{R} & \multicolumn{1}{c}{F1} & \multicolumn{1}{c}{AUC} & \multicolumn{1}{c}{P} & \multicolumn{1}{c}{R} & \multicolumn{1}{c}{F1} & \multicolumn{1}{c}{AUC} & \multicolumn{1}{c}{P} & \multicolumn{1}{c}{R} & \multicolumn{1}{c}{F1} & \multicolumn{1}{c}{AUC} \\ \midrule
SpanOIE                                    & SpanOIE                                           & WiRe57                                        & 9.10                                          & 0.87                  & 0.72                  & 0.788                  & 0.673                   & 0.464                 & 0.194                 & 0.274                  & 0.142                   & 0.464                 & 0.372                 & 0.413                  & 0.272                   \\
\bottomrule
\end{tabular}%
}
\caption{A table that lists performances of different OpenIE systems on the WiRe57 benchmark.}
\end{table*}

\begin{table*}[]
\centering
\resizebox{\textwidth}{!}{%
\begin{tabular}{lllrrrrrrrrrrrrr}
\toprule
\multicolumn{1}{c}{\multirow{2}{*}{Model}} & \multicolumn{1}{c}{\multirow{2}{*}{Training set}} & \multicolumn{1}{c}{\multirow{2}{*}{Test set}} & \multicolumn{1}{c}{\multirow{2}{*}{Sen./Sec}} & \multicolumn{4}{c}{OIE2016}                                                                      & \multicolumn{4}{c}{WiRe57}                                                                       & \multicolumn{4}{c}{CaRB}                                                                         \\ \cmidrule(lr){5-8}  \cmidrule(lr){9-12}  \cmidrule(lr){13-16} 
\multicolumn{1}{c}{}                       & \multicolumn{1}{c}{}                              & \multicolumn{1}{c}{}                          & \multicolumn{1}{c}{}                          & \multicolumn{1}{c}{P} & \multicolumn{1}{c}{R} & \multicolumn{1}{c}{F1} & \multicolumn{1}{c}{AUC} & \multicolumn{1}{c}{P} & \multicolumn{1}{c}{R} & \multicolumn{1}{c}{F1} & \multicolumn{1}{c}{AUC} & \multicolumn{1}{c}{P} & \multicolumn{1}{c}{R} & \multicolumn{1}{c}{F1} & \multicolumn{1}{c}{AUC} \\ \midrule
SpanOIE                                    & SpanOIE                                           & ReOIE2016                                     & 16.87                                         & 0.741                 & 0.842                 & 0.788                  & 0.733                   & 0.772                 & 0.595                 & 0.672                  & 0.527                   & 0.772                 & 0.61                  & 0.681                  & 0.54                    \\
\bottomrule
\end{tabular}%
}
\caption{A table that lists performances of different OpenIE systems on the ReOIE2016 benchmark.}
\end{table*}

\begin{table*}[]
\centering
\resizebox{\textwidth}{!}{%
\begin{tabular}{lllrrrrrrrrrrrrr}
\toprule
\multicolumn{1}{c}{\multirow{2}{*}{Model}} & \multicolumn{1}{c}{\multirow{2}{*}{Training set}} & \multicolumn{1}{c}{\multirow{2}{*}{Test set}} & \multicolumn{1}{c}{\multirow{2}{*}{Sen./Sec}} & \multicolumn{4}{c}{OIE2016}                                                                      & \multicolumn{4}{c}{WiRe57}                                                                       & \multicolumn{4}{c}{CaRB}                                                                         \\ \cmidrule(lr){5-8}  \cmidrule(lr){9-12}  \cmidrule(lr){13-16} 
\multicolumn{1}{c}{}                       & \multicolumn{1}{c}{}                              & \multicolumn{1}{c}{}                          & \multicolumn{1}{c}{}                          & \multicolumn{1}{c}{P} & \multicolumn{1}{c}{R} & \multicolumn{1}{c}{F1} & \multicolumn{1}{c}{AUC} & \multicolumn{1}{c}{P} & \multicolumn{1}{c}{R} & \multicolumn{1}{c}{F1} & \multicolumn{1}{c}{AUC} & \multicolumn{1}{c}{P} & \multicolumn{1}{c}{R} & \multicolumn{1}{c}{F1} & \multicolumn{1}{c}{AUC} \\ \midrule
SpanOIE                                    & SpanOIE                                           & CaRB                                          & 17.14                                         & 0.81                  & 0.778                 & 0.794                  & 0.704                   & 0.609                 & 0.273                 & 0.377                  & 0.219                   & 0.609                 & 0.403                 & 0.485                  & 0.324                   \\
\bottomrule
\end{tabular}%
}
\caption{A table that lists performances of different OpenIE systems on the CaRB benchmark.}
\end{table*}

\begin{table*}[]
\centering
\resizebox{\textwidth}{!}{%
\begin{tabular}{lllrrrrrrrrrrrrr}
\toprule
\multicolumn{1}{c}{\multirow{2}{*}{Model}} & \multicolumn{1}{c}{\multirow{2}{*}{Training set}} & \multicolumn{1}{c}{\multirow{2}{*}{Test set}} & \multicolumn{1}{c}{\multirow{2}{*}{Sen./Sec}} & \multicolumn{4}{c}{OIE2016}                                                                      & \multicolumn{4}{c}{WiRe57}                                                                       & \multicolumn{4}{c}{CaRB}                                                                         \\ \cmidrule(lr){5-8}  \cmidrule(lr){9-12}  \cmidrule(lr){13-16} 
\multicolumn{1}{c}{}                       & \multicolumn{1}{c}{}                              & \multicolumn{1}{c}{}                          & \multicolumn{1}{c}{}                          & \multicolumn{1}{c}{P} & \multicolumn{1}{c}{R} & \multicolumn{1}{c}{F1} & \multicolumn{1}{c}{AUC} & \multicolumn{1}{c}{P} & \multicolumn{1}{c}{R} & \multicolumn{1}{c}{F1} & \multicolumn{1}{c}{AUC} & \multicolumn{1}{c}{P} & \multicolumn{1}{c}{R} & \multicolumn{1}{c}{F1} & \multicolumn{1}{c}{AUC} \\ \midrule
SpanOIE                                    & SpanOIE                                           & LSOIE                                         & 18.56                                         & 0.745                 & 0.851                 & 0.794                  & 0.742                   & 0.537                 & 0.388                 & 0.451                  & 0.298                   & 0.537                 & 0.551                 & 0.544                  & 0.423                   \\
\bottomrule
\end{tabular}%
}
\caption{A table that lists performances of different OpenIE systems on the LSOIE benchmark.}
\vspace{-0.1in}
\end{table*}

\begin{table*}[]
\centering
\resizebox{0.8\textwidth}{!}{%
%
} & non-N-ary model, N-ary test & 0         & 11        & 0    & 4         \\ \cmidrule{2-6} 
                                                                                     & N-ary model, N-ary test     & 4         & 9         & 0             & 2            \\ \bottomrule
\end{tabular}%
}
\caption{Statistical significance tests to answer R1. Each number represents the number of test set and evaluation metric combinations with the corresponding t-score and p-value. When t-score is greater than 0, non-N-ary outperforms N-ary, and when t-score is less than 0, N-ary outperforms non-N-ary.}
\label{table:R1}
\vspace{-0.1in}
\end{table*}

\begin{table*}[]
\centering
\resizebox{0.8\textwidth}{!}{%
\begin{tabular}{cccccc}
\toprule
\multirow{2}{*}{Independent Var.} & \multirow{2}{*}{Constants} & \multicolumn{2}{c}{p-value $\leq 0.05$}     & \multicolumn{2}{c}{p-value $> 0.05$}     \\ \cmidrule(lr){3-4} \cmidrule(lr){5-6} 
                                  &                            & t-score $> 0$ & t-score $< 0$ & t-score $> 0$ & t-score $< 0$ \\ \midrule
                                  \multirow{5}{*}{\begin{tabular}[c]{@{}c@{}}non-IN model \\ vs.\\ IN model\end{tabular}} & non-IN train, IN test         & 9                             & 0                             & 2                                 & 1                                \\ \cmidrule{2-6} 
                                                                                        & IN train, IN test             & 0                             & 4                             & 7                                 & 1                                \\ \cmidrule{2-6} 
                                                                                        & non-IN train, non-IN test     & 0                             & 1                             & 2                                 & 0                                \\ \cmidrule{2-6} 
                                                                                        & IN train, non-IN test         & 3                             & 0                             & 0                                 & 0                                \\ \midrule
\multirow{5}{*}{\begin{tabular}[c]{@{}c@{}}non-IN train \\ vs.\\ IN train\end{tabular}} & non-IN model, IN test         & 6                             & 6                             & 0                                 & 0                                \\ \cmidrule{2-6} 
                                                                                        & IN model, IN test             & 2                             & 7                             & 0                                 & 3                                \\ \cmidrule{2-6} 
                                                                                        & non-IN model, non-IN test     & 2                             & 1                             & 0                                 & 0                                \\ \cmidrule{2-6} 
                                                                                        & IN model, non-IN test         & 2                             & 0                             & 1                                 & 0                                \\ \bottomrule
\end{tabular}
}
\caption{Statistical significance tests to answer R2. Each number represents the number of test set and evaluation metric combinations with the corresponding t-score and p-value. When t-score is greater than 0, non-IN outperforms IN, and when t-score is less than 0, IN outperforms non-IN.}
\label{table:R2}
\vspace{-0.1in}
\end{table*}

\begin{table}[]
\centering
\begin{adjustbox}{width=\linewidth}
\begin{tabular}{lrlrrr}
\toprule
                                                \multicolumn{2}{c}{Configuration 1}                      & \multicolumn{2}{c}{Configuration 2}                      & \multicolumn{1}{c}{\multirow{2}{*}{t-Score}} & \multicolumn{1}{c}{\multirow{2}{*}{p-value}} \\ \cmidrule(lr){1-2} \cmidrule(lr){3-4} 
                                                                                        \multicolumn{1}{c}{Model} & \multicolumn{1}{c}{Sen./Sec} & \multicolumn{1}{c}{Model} & \multicolumn{1}{c}{Sen./Sec} & \multicolumn{1}{c}{}                         & \multicolumn{1}{c}{}                         \\ \midrule
 IMoJIE                    & 2.070                        & Multi$^2$OIE                 & 29.225                       & -21.621                                      & 1.50E-15                                     \\
                                                                                                    IMoJIE                    & 2.070                        & IGL-OIE                   & 84.072                       & -5.501                                       & 2.63E-05                                     \\
                                                                                                    IMoJIE                    & 2.070                        & CIGL-OIE                  & 68.800                       & -4.929                                       & 9.31E-05                                     \\
                                                                            IMoJIE                    & 2.070                        & OpenIE6                   & 28.357                       & -5.813                                       & 1.31E-05                                     \\ \bottomrule
\end{tabular}%
\end{adjustbox}
\caption{Statistical significance tests to answer R3 with \textit{Generative Model vs. Non-generative Model} independent variable . Sentences per second is averaged over all training and test sets.}
\label{table:R3}
\vspace{-0.1in}
\end{table}

\end{document}